\documentclass[10pt,twocolumn,letterpaper]{article}

\usepackage[
    top=0.75in, 
    bottom=1.0in, 
    left=0.625in, 
    right=0.625in, 
    columnsep=0.24in
]{geometry}

\usepackage{amsmath}
\usepackage{lipsum} % Apenas para gerar texto automático de teste
\usepackage{tabularx}
\usepackage{censor}
\usepackage{natbib}
\usepackage{graphicx} % Pacote necessário para inserir imagens
\usepackage{authblk}

\title{JEPAhead: From Inpainting to Causal Forecasting for Urban Driving}
\author[1]{Daniel G. P. Petrini\thanks{daniel.petrini@eldorado.org.br}} 
\author[2]{Davi Pincinato \thanks{d157810@dac.unicamp.br}} 
\author[1]{Fabio Grassiotto\thanks{fabio.grassiotto@eldorado.org.br}} 
\author[1]{Pedro L. F. Prado\thanks{pedro.flores@eldorado.org.br}} 
\author[1]{Alexandre F. S. Osorio\thanks{alexandre.osorio@eldorado.org.br}} 
\author[1]{Raphael V. Ramos\thanks{raphael.ramos@eldorado.org.br}} 
\author[1]{Felipe Pamboukian\thanks{felipe.pamboukian@eldorado.org.br}} 
\affil[1]{Eldorado Research Institute, Campinas, Brazil} 
\affil[2]{University of Campinas (Unicamp), Campinas, Brazil}

\date{}

\begin{document}

\maketitle

\begin{abstract}

%Forecasting driving-relevant representations offers an alternative to pixel-space video prediction. 
We fine-tune a pre-trained V-JEPA predictor from bidirectional latent inpainting to action-free causal forecasting while keeping the encoder frozen. Using contiguous past observations, horizon-aligned targets, and group-level causal attention, the model forecasts eight future token groups, corresponding to sixteen frames, in one forward pass. On CARLA Town12, it retrieves the correct future among 591 candidates in 69.0\% of cases, compared with 55.3\% for persistence, with greater gains at longer horizons and on tokens exhibiting greater change. It also outperforms a compact predictor trained from scratch with six times fewer parameters. Semantic analysis indicates driving-relevant information in predicted features: a fixed linear probe trained on real encoder tokens achieves 0.16 mIoU on patches undergoing semantic-class changes, versus 0.13 for persistence. Vehicle-conditioned retrieval improves over persistence by 15 percentage points at the longest horizon. Pixel-space evaluation shows modest L1 gains but little improvement under LPIPS, motivating direct evaluation through latent and semantic measures alongside an oracle reconstruction reference. These findings support further investigation of latent forecasting for driving perception, while downstream utility remains unverified.

\end{abstract}

\section{Introduction}

\subsection{Motivation}

World models are learned models that represent an environment and predict how it evolves over time. In autonomous driving, they can use past sensor observations to anticipate traffic evolution and, when conditioned on proposed ego actions, predict the outcomes of alternative driving decisions \cite{hu2023gaia1}. Future prediction also provides a learning signal for scene representations, connecting world modeling with the development of features useful for driving perception and decision-making \cite{li2025law}.

Motivated by this potential, this paper introduces an adaptation of the original V-JEPA (Visual Joint Embedding Predictive Architecture) \cite{bardes2024vjepa}, framework tailored for the automotive context. Distinct from expensive conventional generative models, our approach bypasses pixel decoding entirely, performing future-horizon prediction directly via tokens within the latent space (a separately trained decoder is used only for pixel-space evaluation). By discretizing and predicting the temporal evolution of the scene at the embedding level, the model focuses strictly on the structural semantics of the environment (driver scenes). 

Our results demonstrate that this formulation not only mitigates computational overhead but also optimizes the representation of spatio-temporal uncertainty in traffic scenarios. 
%This work lays the foundation  for highly efficient tokenized world models viable for real-time embedded systems.
We also provide an initial step toward compact token-based world models, with potential for future use in resource-constrained systems.

The remainder of this section reviews related work and presents our contributions. We then describe the methodology, report the results, and discuss their implications. The paper concludes with the main findings and directions for future research.

%[Versao 2]
%Anticipating how a scene will evolve is a prerequisite for acting in it. In autonomous driving this is usually approached by generating future sensor observations, or by supervising intermediate representations with annotated perception targets. Both routes are expensive: the first commits substantial capacity to rendering pixels that the controller never consumes, and the second requires labels that do not scale.

%A third route predicts in REPRESENTATION SPACE. If a self-supervised encoder already maps video to features that capture scene structure, a predictor can be trained to forecast those features directly, never reconstructing observations. The appeal is economy: the predictor operates on a few thousand tokens rather than millions of pixels, and needs no annotation.

%The appeal is also the difficulty. A representation that is never decoded offers no direct way to verify that a forecast is correct, and the metrics customarily used to check such systems are defined on pixels. This paper studies both halves of that proposition: whether a frozen V-JEPA1 encoder supports causal forecasting, and how such a claim can be evaluated without being misled by the decoder.

\subsection{Related Work}

\subsubsection{Joint-embedding predictive architectures}

These models learn to predict target representations from context, with the prediction loss computed in embedding space.
I-JEPA \cite{assran2023ijepa} applies this principle to masked image regions, while V-JEPA \cite{bardes2024vjepa} extends it to spatiotemporal regions in video without pixel reconstruction or contrastive negatives. V-JEPA 2-AC \cite{assran2025vjepa2} scales video pre-training and introduces an action-conditioned world model through additional training on robot interaction data, enabling planning in representation space.

Our work examines whether a V-JEPA predictor pre-trained for bidirectional latent inpainting can be adapted to causal future forecasting. V-JEPA provides a starting point for studying this transfer, whereas V-JEPA 2-AC already incorporates action-conditioned predictive training.

\subsubsection{World models and latent dynamics}

World Models \cite{ha2018worldmodels} combines a variational autoencoder with recurrent latent dynamics, supporting controller learning within imagined environments. Dreamer \cite{hafner2023dreamerv3} develops this approach by learning policies through latent rollouts, with representations shaped partly by observation reconstruction. For autonomous driving, MILE \cite{hu2022mile} jointly learns latent dynamics and a driving policy using geometric structure and bird’s-eye-view semantic prediction, while GAIA-1 \cite{hu2023gaia1} combines autoregressive token prediction with video diffusion decoding to generate driving scenarios.

These approaches use observation reconstruction or generation within their learning pipelines. Our work instead studies the transfer of a predictor pre-trained solely through feature prediction to causal forecasting with a frozen encoder. A separately trained decoder supports visual evaluation without contributing to predictor training.

\subsubsection{Latent prediction and representation analysis for driving}

Feature-space prediction has been explored for driving and visual control. LAW \cite{li2025law} predicts future scene features conditioned on ego actions, while AD-L-JEPA \cite{zhu2025adljepa} learns LiDAR representations through masked spatial prediction. Zhu and Choromanska \cite{zhu2026occupancy} extend JEPA-based learning to spatiotemporal LiDAR prediction, evaluated through occupancy completion and forecasting. Drive-JEPA \cite{wang2026drivejepa} combines video JEPA pre-training with trajectory planning. Complementing these approaches, Jeon et al. \cite{jeon2026probing} use linear probes and targeted interventions to examine predictive information in driving policies. Beyond driving, DINO-WM \cite{bar2024dinowm} learns action-conditioned dynamics over frozen DINOv2 patch features for visual planning.

These studies motivate prediction and analysis in learned feature spaces. Our work specifically examines adapting a pre-trained bidirectional V-JEPA predictor to action-free causal forecasting. % while keeping the encoder frozen.

\subsection{Contributions}

We study causal forecasting in the representation space of a frozen V-JEPA encoder on simulated urban driving, and report the following.

\begin{enumerate}

\item \textit{A causal adaptation of the pre-trained V-JEPA predictor}. We initialize the predictor from pre-trained weights and fine-tune it using contiguous past-token context, future-position queries, and a group-level causal attention mask, while keeping the encoder frozen. It predicts eight future groups in one forward pass. A compact predictor trained from scratch provides a comparison with approximately six times fewer parameters.

\item \textit{Evidence of sequence-specific forecasting in representation space.} Predicted tokens retrieve the correct future among 591 candidates in 69.0\% of cases, compared with 55.3\% for persistence, while retaining 98.4\% of the target population’s spread. The advantage over persistence grows with the forecast horizon and is larger on tokens exhibiting greater change.

\item \textit{Evidence of driving-relevant information in predicted features.} A fixed linear probe trained on real encoder tokens achieves 0.16 mIoU on patches undergoing semantic-class changes, compared with 0.13 for persistence. Complementary class-conditioned retrieval shows improvements over persistence across all six evaluated classes at the longest horizon.

\item \textit{A characterization of limited gains in pixel space.} Improvements in latent forecasting yield modest L1 gains but little improvement under LPIPS with the evaluated decoder. We report reconstructions from true future tokens as an oracle reference and normalize pixel skill by the oracle-referenced achievable reduction to contextualize these results.

\end{enumerate}

%We are explicit about what these results do not establish. They come from one simulated domain and a horizon of roughly 16 frames or 8 groups; the predictor is action-free, so it forecasts but cannot answer counterfactual queries; and persistence is a strong baseline whose value we report alongside our own throughout, since a substantial share of moment identification is already available in the last observed token.

Evaluation covers CARLA PDM-Lite \cite{beisswenger2024pdmlite} Town12 and 16 future frames (eight temporal groups); generalization and action-conditioned counterfactual prediction remain untested.

%unconditional => action-free

\section{Methodology}

\subsection{Problem formulation} \label{problem_formulation}

We study causal future prediction in the representation space of a frozen video encoder. Following the joint-embedding predictive architecture (JEPA) formulation, a predictor maps a context representation $s_x = Enc(x)$  to an estimate of a target representation $s_y = Enc(y)$, conditioned on a variable z that specifies what is to be predicted:

%\begin{equation}
%    s_{y_{hat}} = Pred(s_x, z),        L = d(s_{y_{hat}}, s_y)
%\end{equation}

\begin{equation}
\hat{s}_y = Pred(s_x, z), \quad L = d(\hat{s}_y, s_y)
\end{equation}

In V-JEPA pre-training, $x$ is a sparse set of visible patches, $y$ is a set of
masked patches drawn from the same temporal window, and $z$ encodes the position
of the masked region. The task is therefore spatio-temporal inpainting: $y$ may
lie before, during, or after $x$ in time.

\begin{figure}[htbp]
    \centering
    \includegraphics[width=0.45 \textwidth]{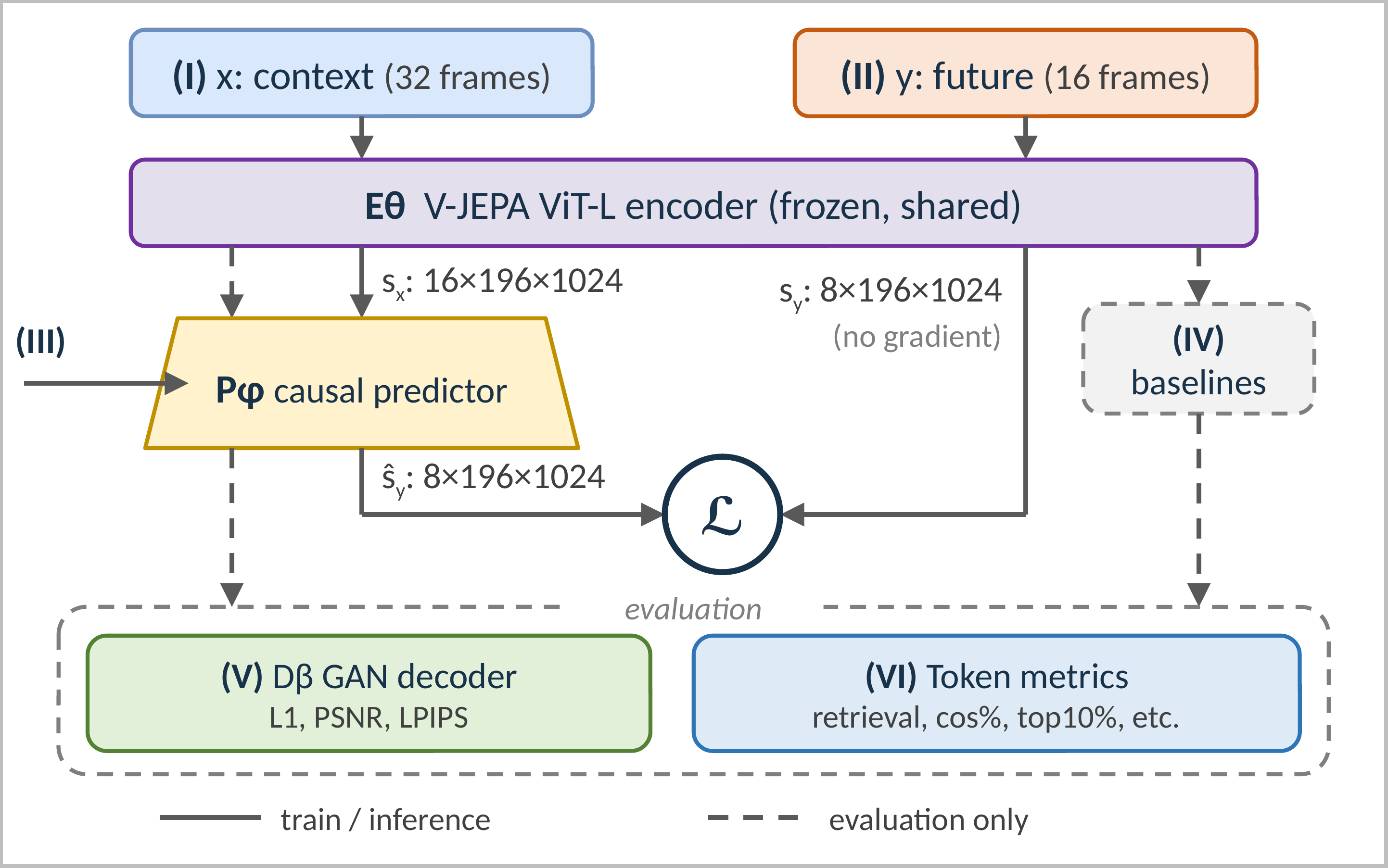}
    \caption{Causal future prediction in the frozen V-JEPA latent space (adapted from \cite{assran2023ijepa}). %Context x and target y are encoded by the same frozen encoder Eθ; 
Only the predictor P$\varphi$ is trained, with latent loss $\mathcal{L}$ between ŝy and sy.
(I) Context: 32 frames (16 temporal groups) of a 48-frame CARLA clip at 4 Hz.
(II) Future target: 16 frames (8 groups).%; target tokens receive no gradient.
(III) $z$: 8 future queries (mask tokens + 3D positional embedding). %; causal mask lets each future step see the full context and only earlier horizons.
(IV) Evaluation baselines computed from the encoder tokens.
(V) Predicted tokens decoded by GAN decoder D$\beta$, for image metrics.
(VI) Token-space metrics.}
    \label{fig:exemplo}
\end{figure}

We retain this formulation but change the temporal relation between $x$ and $y$.
In our setting $x$ is a contiguous block of past observations, $y$ is strictly
posterior to $x$, and $z$ addresses future instants that lie outside the observed
window, as can be seen in Figure \ref{fig:exemplo}. The task becomes causal forecasting. Crucially, $z$ remains available at
inference time (it is a positional address, not content), whereas $y$ is never
provided to the model.

\subsection{Frozen encoder and tokenization}

We use a pre-trained V-JEPA ViT-L/16 encoder (24 blocks, embedding dimension 1024, 16 heads) and keep it frozen throughout. Input clips are 32 RGB frames at 224 $\times$ 224. A 3D patch embedding with kernel and stride (2, 16, 16) maps each tubelet of two frames and 16 $\times$ 16 pixels to one token, yielding 16 temporal positions  $\times$  14 $\times$ 14 spatial positions = 3136 tokens per clip. A 3D sine-cosine positional embedding is added, and 24 transformer blocks are applied with dense, unmasked attention.

Because the flattening order is temporal-major, the token sequence can be read
as 16 consecutive groups of 196 tokens, where group $k$ corresponds to frames
$(2k, 2k+1)$. We use "group" throughout to denote this slice. %; it is a reading convention, not a structural unit of the encoder. We emphasize that encoder attention is dense and bidirectional: a token of group k is contextualized by the entire 32-frame clip, a property with direct consequences for how targets must be constructed.

%For target construction, true future in latent space, we encode ONE WINDOW PER HORIZON, each ending on its own target group. For horizon h in $1..H$, the window spans clip frames $[2h, 2h + 32)$ and we retain only its LAST group, ensuring that every target is causally encode. 

%This has two consequences. First, every target occupies the same window position as the last observed context group, so the positional signature is constant across horizons. Second, every target is causally encoded: it has seen only its own 32 frames of past and nothing after. In the notation of Section \ref{problem_formulation}, the alignment makes $y$ consistent with what $z$ addresses.

To construct ground-truth future representations, we encode one 32-frame window per horizon, each ending at its target group. For horizon $h \in \{1,\ldots,H\}$, the window spans clip frames $[2h, 2h+32)$, and we retain only its last temporal group. This ensures that each target is computed without access to frames beyond its target group.

\subsection{Dataset and sample construction}

Our experiments use CARLA driving sequences. We used the PDM-Lite rule-based expert \cite{beisswenger2024pdmlite} to drive Carla Town12 entirely, its routes are partitioned at the route level with seed 42 into 70\% train, 10\% validation and 20\% test, giving 1218 /174 / 348 routes. Route-level splitting prevents frames of the same route from appearing in more than one split.

%Each sample is a 48-frame clip: 32 context frames and 16 future frames. 
Figure \ref{fig:exemplo} contains frames flows. The sampling window advances 16 frames, producing 6368 / 931 / 1773 samples for train / validation / test. In group units, each sample provides $K = 16$ context groups and $H = 8$ future groups.
Global retrieval uses every third test sample, yielding 591 samples from 301 routes, with all 591 corresponding futures forming the candidate pool.

\subsection{Causal predictor}

We adapt the pre-trained V-JEPA predictor (12
blocks, embedding dimension 384, 16 heads; 23.9M parameters). % without changing its weights at initialization. 
The $K \times 196 = 3136$ context tokens are projected
to the predictor dimension and summed with positional embeddings. $H \times 196 =
1568$ query tokens are formed from a single pre-trained mask token replicated
and summed with the positional embeddings of the future instants; they carry
no content. The two are concatenated into one sequence of 4704 tokens.

Causality is imposed by an attention mask built at group granularity and
expanded by the group size, divided by context and future. %, as in Figure \ref{fig:exemplo}.

%\begin{align}
%\text{context} \times \text{context} &: \text{full (all observed; no leakage possible)} \nonumber \\
%\text{future} \times \text{context} &: \text{full} \nonumber \\
%\text{future} \times \text{future} &: \text{lower triangular} \nonumber \\
%\text{context} \times \text{future} &: \text{blocked}
%\end{align}

Within any permitted pair of groups, attention is unrestricted; the constraint
is purely temporal. Since the reference implementation ignores the mask
argument of its attention module, we reimplement the block forward pass to
supply an explicit additive mask, and select a memory-efficient attention
backend to avoid materializing the dense $4704 \times 4704$ matrix.

The positional embedding is regenerated for $K + H = 24$ groups rather than
loaded from the checkpoint, since the pre-trained predictor was trained for 16
groups and has no representation for instants beyond the observed window.

The output head predicts a residual over the last observed group,
$\hat{s}_y = s_x[K-1] + \Delta$, so that the network models change rather than
absolute state.

\textbf{Compact predictor}. As an efficiency baseline we train a compact predictor (3.97M parameters, 6.0x fewer) that factorizes space and time: features are projected to 256 dimensions, mixed spatially by depthwise $3 \times 3$ convolutions on the $14 \times 14$ grid, then attended temporally as 196 independent sequences of length $K + H$ under the same causal mask, and mixed spatially again before a linear head. It also predicts a residual over the last observed group.

\subsection{Decoder}

The GAN decoder used in the experiments receives a grid of tokens ($14 \times 14 \times 1024$) and generates two RGB frames of ($224 \times 224$). Its generator has 6.97 million parameters and includes six residual blocks, a spatial self-attention module, four PixelShuffle magnification stages, and two output heads (21 convolutions in total). During training, it uses a PatchGAN discriminator with 2.76 million parameters, which is dispensed in inference.

\subsection{Training}

%Both predictors are trained on cached frozen-encoder tokens, so the encoder is never run during training. The loss combines cosine and Smooth-L1 terms on the predicted tokens with three terms on temporal differences taken with respect to the last observed group:

Both predictors are trained on cached tokens from a frozen encoder, so the encoder is never run during predictor training. The objective combines cosine and Smooth-L1 losses on predicted tokens, Smooth-L1 and cosine losses on consecutive-group temporal differences initialized with the last observed group, and a Smooth-L1 loss on second-order temporal differences:

%Both compare predicted and ground-truth changes between consecutive temporal groups:
%- \(\mathcal{L}_{\mathrm{delta}}\): Smooth-L1 loss on the change vectors, penalizing errors in their magnitude and components.
%- \(\mathcal{L}_{\mathrm{deltadir}}\): cosine loss on those same vectors, penalizing errors in their direction.
%For the first future group, the change is measured from the last observed group; subsequent changes are measured from the preceding future group.

%\begin{equation}
%\begin{split}
%    L ={} & 1.0 * L_{cos} + 0.5 * L_{sl1} + 0.5 * L_{delta} \\
%        & + 0.2 * L_{delta_{dir}} + 0.1 * L_{accel}
%\end{split}
%\end{equation}

\begin{align}
L = {} & 1.0 \cdot L_{cos} + 0.5 \cdot L_{sl1} + 0.5 \cdot L_{delta} \nonumber \\
       & + 0.2 \cdot L_{deltadir} + 0.1 \cdot L_{accel}
\end{align}

For training parameters, after exploring parameters we converged to AdamW, batch size 32 without gradient accumulation, weight decay 1e-4, gradient clipping 1.0, fp16 mixed precision, 3 epochs of linear warm-up followed by cosine annealing to 1e-6, early stopping with patience 10. Peak learning rate 6e-4 for the adapted predictor (30 epochs) and 1.5e-3 for the compact predictor (50 epochs). Both runs stopped early, with the best epoch reached at 53\% and 35\% of peak learning rate respectively.

%; both values were selected by a prior sweep in which higher rates degraded validation loss. Both runs stopped early, with the best epoch reached at 53\% and 35\% of peak learning rate respectively and no improvement during the subsequent annealing phase.

\subsection{Baselines}

Three analytic baselines are computed from the context tokens with no
training. All are element-wise and treat each of the 196 spatial positions
independently. With $g_k$ the $k_{th}$ context group and h the horizon:

\begin{equation}
\begin{split}
    & \text{persistence} : g_{K-1} \\
    & \text{linear}      : g_{K-1} + h (g_{K-1} - g_{K-2}) \\
    & \text{EMA}         : g_{K-1} + h \cdot \sum_i w_i (g_{i+1} - g_i),  w_i \propto decay^{(K-2-i)}
\end{split}
\end{equation}

Linear and EMA are the same constant-velocity extrapolation and differ only in
how velocity is estimated. Persistence, the last group frozen through horizon, is the reference against which
forecasting skill is measured.

\subsection{Evaluation protocol}

\textbf{Latent metrics}. To explore the performance on latent space we used metrics as Per-token cosine similarity between predicted and true future tokens, reported globally and per horizon. Because a driving scene is largely static, we additionally report cosine restricted to the top 10\% and top 25\% of tokens ranked by $||y - s_x[K-1]||$, i.e. those that actually moved.

\textbf{Distributional metrics}. Over the evaluation population we report (i) batch
retrieval, ranking every sample's true future against all candidates by cosine
and measuring top-1 and top-5 accuracy; (ii) a collapse ratio, the standard
deviation of predictions across samples divided by that of the true futures;
and (iii) a Frechet distance between predicted and true token distributions,
computed in a shared PCA basis because the sample count is small relative to
the token dimension.

% Retirando Hard negatives - 2AFC comparações - nao muito importante e nao tem espaço.
%\textbf{Hard-negative control}. Batch retrieval draws its distractors from all routes and therefore cannot separate identifying the MOMENT from recognizing the SCENE. We add a two-alternative forced choice restricted to a single route: for every pair $(i, j)$ of samples from the same route, the model scores a hit when $\cos(\hat{s}_{y_i}, s_{y_i}) > \cos(\hat{s}_{y_i}, s_{y_j})$. Chance is 0.5. Accuracy is reported as a function of the temporal separation between the two samples; pairs closer than one clip length share raw frames, so the headline figure uses only pairs separated by at least 48 frames.

\textbf{Pixel and perceptual metrics}. Predicted tokens are decoded to frames by a
GAN decoder fine-tuned on the same cached tokens, and compared to ground truth
by L1, PSNR and LPIPS \cite{zhang2018lpips}. We also decode the TRUE future tokens, giving an oracle
that bounds what any predictor can achieve through this decoder. LPIPS uses an
AlexNet backbone: the decoder is trained with a VGG-16 perceptual loss, so a
VGG-based LPIPS would evaluate it against a close relative of its own training
objective.

\textbf{Skill scores}. For an error metric $e$, forecasting skill relative to persistence is $1 - e_{model} / e_{persistence}$. For pixel metrics we additionally report the
fraction of the achievable reduction that is captured,

\begin{equation}
\label{eq:skill}
\frac{e_{persistence} - e_{model}}{e_{persistence} - e_{oracle}}
\end{equation}

which separates predictor limitations from decoder limitations. This
normalization is meaningful only in pixel space; in latent space the oracle
error is zero by construction and the expression reduces to the ordinary skill
score.

\textbf{Experimental controls}. Evaluation uses a stride that makes consecutive evaluated samples temporally disjoint, so that near-duplicate windows do not enter the retrieval pool.

%Levamos a parte mais importante para a subseção acima

%\subsection{Experimental controls}

%All comparisons hold constant: the frozen encoder and its checkpoint, the route partition and sample ordering, the token cache, the decoder, the optimization regime (batch size, effective batch, learning-rate schedule, weight decay, precision) and the output parameterization (both predictors predict residuals). Evaluation uses a stride that makes consecutive evaluated samples temporally disjoint, so that near-duplicate windows do not enter the retrieval pool.

%[NOTE] If a "threats to validity" paragraph is included, the confounds
%controlled during this investigation, in order, were: context window, decoder calibration, output parameterization, optimization regime, hyper-parameter search, learning-rate ceiling, schedule, target alignment, shard size (which sets batch composition, since batches never span shards) and encoder batch size (numerical reproducibility).

\section{Results}

\subsection{Forecasting in representation space}

\begin{table}[tbp]
\caption{Reports the main comparison on the test split. $\uparrow$/$\downarrow$ indicate higher/lower is better. For standard ratio, ideal:1}
\label{tab:main_comparison}
\centering
\footnotesize
\setlength{\tabcolsep}{3pt} 
%\begin{tabular}{lccccccc}
\begin{tabularx}{\columnwidth}{lXXXXXXX}
\hline
\textbf{Method} & \textbf{cos $\uparrow$} & \textbf{top25 (\%) $\uparrow$}  & \textbf{top10 (\%) $\uparrow$} & \textbf{retr. @1 $\uparrow$} & \textbf{retr. @5 $\uparrow$}  & \textbf{std ratio } & \textbf{Frechet $\downarrow$} \\ \hline
  original    & 0.8478   & 0.7113   & 0.6508  & 0.6904  & 0.8479  & 0.984    & 5.95 \\
  compact     & 0.8421   & 0.7021   & 0.6414  & 0.5882  & 0.7728  & 0.934    & 5.28 \\
  persistence & 0.7814   & 0.5309   & 0.4249  & 0.5529  & 0.7418  & 1.018    & 2.52 \\
  EMA         & 0.6609   & 0.3573   & 0.2399  & 0.3780  & 0.5909  & 1.603   & 90.12 \\
  linear      & 0.4920   & 0.2161   & 0.1202  & 0.2481  & 0.4473  & 3.047  & 632.27 \\
  chance      &    —     &  —       &  —  & 0.00169    & —   & —     &  —    \\ \hline
\end{tabularx}
\end{table}

Both learned predictors outperform the analytic baselines in cosine similarity and retrieval (Table 1). The adapted predictor achieves the highest global cosine similarity (0.8478 versus 0.8421 for the compact model and 0.7814 for persistence), corresponding to 30.4\% skill relative to persistence. Its advantage increases on dynamic tokens, reaching 39.3\% skill on the top 10\%. Retrieval more clearly separates the learned models: the adapted predictor achieves 69.0\% top-1 and 84.8\% top-5 accuracy, compared with 58.8\% and 77.3\% for the compact model and 55.3\% and 74.2\% for persistence. EMA and linear extrapolation underperform persistence across all these measures.

We additionally evaluated within-route temporal discrimination using two-alternative forced choice over 11,348 pairs separated by at least 48 frames. Accuracy was 91.9\%, compared with 87.4\% for persistence (chance: 50\%). Because routes in CARLA Town12 share similar scenery, route restriction does not necessarily provide harder distractors than global retrieval.

\subsection{Per-horizon behaviour}

Retrieval accuracy decays with the horizon for every method, but not at the
same rate, as seen in Table \ref{tab:horizon}.

\begin{table}[tbp]
\caption{Horizon of predictions}
\label{tab:horizon}
\centering
\footnotesize
\setlength{\tabcolsep}{3pt} 
\begin{tabular}{lcccccccc}
\hline
\textbf{h}  & \textbf{1} & \textbf{2} & \textbf{3} & \textbf{4} & \textbf{5} & \textbf{6} & \textbf{7} & \textbf{8} \\ \hline
  original     & 0.914  & 0.829  & 0.739  & 0.709  & 0.638  & 0.602  & 0.579  & 0.513 \\
  compact      & 0.848  & 0.753  & 0.645  & 0.604  & 0.548  & 0.462  & 0.443  & 0.403 \\
  persistence  & 0.858  & 0.714  & 0.623  & 0.563  & 0.492  & 0.415  & 0.399  & 0.359 \\
  ratio o/p    & 1.07   & 1.16   & 1.19   & 1.26   & 1.30   & 1.45   & 1.45   & 1.43 \\ \hline
\end{tabular}
\end{table}

At h = 1 the adapted predictor barely exceeds trivial continuation; by
h = 6..8 it exceeds it by roughly 45\%. The advantage of prediction grows
precisely where continuation fails.

\subsection{Distributional properties}

Also looking to Table \ref{tab:main_comparison}, the adapted and compact predictors retain population variability close to that of the targets, with standard-deviation ratios of 0.984 and 0.934, respectively, providing evidence against population-level collapse. Their Fréchet distances are also substantially lower than those of EMA and linear extrapolation (5.95 and 5.28 versus 90.12 and 632.27), whose inflated variability indicates unstable extrapolation in feature space. Persistence nevertheless achieves the lowest Fréchet distance (2.52), consistent with reusing genuine encoder features. Thus, the learned predictors improve future-specific accuracy without surpassing persistence in distributional agreement.

\subsection{Pixel and perceptual evaluation} \label{pixel_eval}

\begin{table}[tbp]
\caption{Pixel and Perceptual evaluation}
\label{tab:pixel_eval}
\centering
\footnotesize
\setlength{\tabcolsep}{3pt} 
\begin{tabular}{lccccc}
\hline
\textbf{method} & \textbf{L1 $\downarrow$} & \textbf{top10\% $\downarrow$} & \textbf{PSNR $\uparrow$} & \textbf{LPIPS $\downarrow$} & \textbf{LPIPS top10\% $\downarrow$} \\ \hline
  original     & 0.0612   & 0.1594   & 21.01   & 0.2926     & 0.3784 \\
  compact      & 0.0621   & 0.1656   & 20.89   & 0.2916     & 0.3802 \\
  persistence  & 0.0636   & 0.1843   & 20.63   & 0.2907     & 0.3851 \\
  EMA          & 0.0801   & 0.2116   & 18.92   & 0.3386     & 0.4245 \\
  linear       & 0.1297   & 0.2303   & 15.26   & 0.5157     & 0.5519 \\
  oracle       & 0.0443   & 0.0869   & 23.51   & 0.2269     & 0.2711 \\ \hline
\end{tabular}
\end{table}

%Decoding the true future tokens yields L1 0.0443 and PSNR 23.51 dB (Table \ref{tab:pixel_eval}), demonstrating substantial reconstruction error even without forecasting error. Under L1, the adapted predictor improves on persistence by approximately 3.8\% globally and 13.5\% over dynamic pixels, capturing 25.6\% of the oracle-referenced achievable reduction on dynamic pixels. Under LPIPS, it slightly underperforms persistence globally, with only a 1.7\% improvement over dynamic pixels.
%Spatial gradient energy provides a possible explanation for this discrepancy. Relative to ground truth, it decreases by 37.1\% for the oracle, 40.5\% for persistence, and 43.6\% and 43.2\% for the adapted and compact predictors, respectively. Substantial smoothing therefore remains even with true future tokens, while predicted tokens introduce additional smoothing. This is consistent with the modest L1 gains failing to translate into comparable perceptual improvements.

Decoding true future tokens yields L1 0.0443 and PSNR 23.51 dB, revealing reconstruction error even without forecasting error (Table 3). The adapted predictor reduces L1 relative to persistence by 3.8\% globally and 13.5\% over dynamic pixels, capturing 25.6\% of the oracle-referenced L1 headroom on dynamic pixels. However, LPIPS slightly worsens globally and improves by only 1.7\% over dynamic pixels. Thus, latent forecasting gains translate into modest pixel-space improvements, while the relative contributions of forecasting and reconstruction limitations to perceptual quality remain unresolved.

%Decoding the TRUE future tokens yields L1 0.0443 and 23.51 dB, as can be seen on Table \ref{tab:pixel_eval}. More than half of the pixel error of any predictor is therefore produced by the decoderalone, independently of forecasting quality.
%Under L1 the adapted predictor improves on persistence by a skill of 0.037 globally and 0.135 over dynamic pixels, capturing 25.6\% of the reduction the oracle shows to be achievable. Under LPIPS the advantage disappears: skill is -0.006 globally and +0.017 over dynamic pixels.
%Measuring spatial gradient energy of the decoded frames explains the discrepancy. Relative to ground truth, the oracle loses 37.1\% of sharpness, persistence 40.5\%, and the learned predictors 43.2\% and 43.6\%. Blur is dominated by the decoder, but the predictors add a further 3 points over persistence, which decodes real tokens. L1 rewards the averaging that produces that blur and LPIPS penalizes it; the two effects cancel.

% Abaixo o impacto do retreino do janelamento-FIX, não relevante ficar comentando toda hora.
%We also observe that replacing the target construction, retraining both
%predictors and re-fine-tuning the decoder changed pixel L1 from 0.0616 to
%0.0612 and the oracle from 0.0444 to 0.0443. A structural change to the
%definition of the prediction target left pixel-aligned metrics unmoved.

\subsection{Capacity}

The adapted V-JEPA predictor leads the compact predictor (6.0x fewer parameters) on every paired metric, most clearly on retrieval (0.690 vs 0.588). The gap in plain cosine is small (0.848 vs 0.842), so the discriminative measures separate the two models where the averaged similarity does not. The compact predictor remains attractive when capacity is constrained, like embedded systems.
%, but its margin over persistence on the hard-negative control is only +0.012, against +0.045 for the adapted predictor.

\section{Discussion}

%\textbf{What the representation supports.} The evidence that a frozen V-JEPA1 encoder carries usable predictive structure does not rest on beating a baseline. It rests on identifiability: predicted tokens retrieve the correct future among 591 candidates 69.0\% of the time, 408x chance. Predictions neither collapse (98.4\% of the true spread) nor leave the manifold (Frechet 5.95 against 632 for linear extrapolation). This is a statement about the representation, and it is independent of how strong any particular baseline happens to be.

\textbf{Predictive value of frozen representations.}
Our results indicate that V-JEPA features contain predictive structure that can be exploited for causal forecasting without updating the encoder. The strongest evidence is future retrieval: predicted tokens identify the corresponding future among 591 candidates in 69.0\% of cases, versus 55.3\% for persistence. Distributional results complement this finding: predictions retain 98.4\% of the target population's standard deviation, with a Fréchet distance of 5.95 versus 632 for linear extrapolation. These findings support the use of frozen features for future prediction, while leaving their utility for downstream driving tasks to be established.

\textbf{Where prediction helps.} Two independent measures show the same structure: the advantage of a learned
predictor over trivial continuation grows with the difficulty of the
discrimination. It grows with the horizon (retrieval ratio 1.07 at h = 1 to
1.43 at h = 8), it is larger on the tokens that actually moved (cosine 0.651
vs 0.425, against 0.848 vs 0.781 overall).

Averaged metrics understate this because driving scenes are dominated by
static content, for which repeating the last observation is very nearly
correct. Reporting only global cosine or global L1 in this domain will
systematically understate the value of prediction.

\textbf{Where forecasting adds value.}
The predictor's advantage over persistence becomes more pronounced at longer horizons and on tokens exhibiting greater feature change. The ratio of predictor to persistence retrieval accuracy rises from 1.07 at $h=1$ to 1.43 at $h=8$. On the 10\% most-changing tokens, the cosine advantage over persistence is 0.226 (0.651 versus 0.425), compared with 0.066 overall. These complementary patterns suggest that global averages can dilute forecasting gains by combining changing regions with relatively stable content, where persistence remains competitive. This supports reporting horizon-specific and change-conditioned results alongside global cosine and pixel-space metrics.

\textbf{Persistence is a strong baseline.} %As we mentioned above, 
Persistence reaches %0.874 on the same-route control and 
0.553 on batch retrieval. A large fraction of what looks like moment identification is
already available in the last observed token, with no prediction at all. The
contribution of the predictor is the increment, and we report both figures
throughout rather than the model's value alone.

%This has a methodological consequence beyond our setting. In the unaligned
%target construction, persistence retrieved at 0.191 and the contrast with the
%model appeared to be 3.1x; with targets correctly aligned the contrast is
%1.25x. A weak-looking baseline should prompt a check of whether the baseline
%is genuinely weak or the evaluation is unintentionally penalizing it.

\label{decoder_discussion}
 
\textbf{ The gain does not reach pixels.} %Improvements in latent forecasting translate into modest L1 gains but little improvement under LPIPS. Even with true future tokens, reconstructions lose 37.1\% of spatial gradient energy relative to ground truth and achieve a PSNR of 23.51 dB, indicating that the
As mentioned in Section \ref{pixel_eval}, encoder–decoder pipeline does not fully preserve image detail. Reconstructions from predicted tokens yield additional smoothing, consistent with L1 improvements that do not translate into comparable perceptual gains. Increasing motion-weighted reconstruction loss also failed to improve decoder performance. These findings motivate alternative decoding approaches, but do not establish whether reconstruction limitations or forecasting errors primarily constrain perceptual quality.

\textbf{Driving-relevant information in predicted features.} 
Predicted features retain semantic information despite limited pixel reconstruction. A nine-class linear probe is trained on real, L2-normalized validation tokens; its weights and validation-fitted standardization remain fixed for route-disjoint testing. mIoU averages six classes with $\ge$1\% validation frequency. On patches undergoing semantic-class changes, predictions achieve 0.16 mIoU, versus 0.13 for persistence and 0.36 for real future tokens.

\begin{figure}[htbp]
    \centering
    \includegraphics[width=0.5 \textwidth]{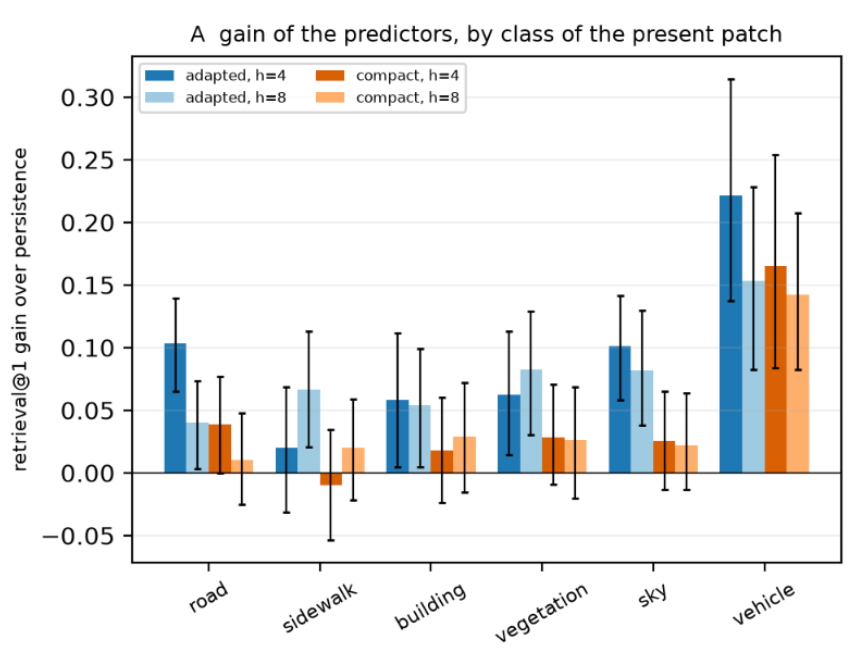}
    \caption{Linear-probe IoU and class-conditioned retrieval@1 gains over persistence at h=4,8. Retrieval intervals show 95\% route-bootstrap confidence intervals (1,000 resamples; seed-42 predictors).}
    \label{fig:predictor_gain_by_class}
\end{figure}

For class-conditioned cosine retrieval, predicted and real future tokens are mean-pooled over each sample’s own patches, selected by majority CARLA labels in the last context frame. Galleries include samples with $\ge$8 class patches; 24 ego-hood patches are excluded throughout. Gains over persistence are positive across all six classes at h=8, with 95\% route-bootstrap intervals excluding zero. Vehicle-associated gains are the largest observed: +0.22 at h=4 and +0.15 at h=8  (Fig. \ref{fig:predictor_gain_by_class}), although unequal gallery sizes (176–572 samples) limit comparisons across classes. These findings motivate further investigation of latent forecasting for driving perception.

%Weak pixel reconstruction does not preclude useful semantic information in predicted features. A patch-wise linear probe trained solely on real encoder tokens and applied unchanged to predictions achieves 0.16 mIoU on patches undergoing semantic-class changes, compared with 0.13 for persistence and 0.36 for real future tokens. Complementary class-conditioned retrieval (Fig. \ref{fig:predictor_gain_by_class}) shows positive gains over persistence across all six evaluated classes at h=8, with 95\% route-bootstrap intervals excluding zero. Vehicle-associated features exhibit the largest observed gains: +0.22 at h=4 and +0.15 at h=8. These results support further investigation of latent forecasting for driving perception, although class-dependent retrieval galleries limit comparisons across categories.

\textbf{Limitations.} Evaluation is restricted to CARLA Town12 and a 16-frame horizon. Generalization to real-world driving and longer horizons remains untested. The predictor is conditioned on past observations and target positions, but not on future ego actions, limiting its use for evaluating alternative action sequences in planning. Although semantic probing and class-conditioned retrieval indicate driving-relevant information, object-level forecasting and downstream utility remain unverified. Pixel-space conclusions depend on decoder quality, while LPIPS and the decoder’s perceptual training loss share ImageNet-based features, potentially affecting evaluation. 

\section{Conclusion}

We show that an inpainting-pre-trained V-JEPA predictor can be fine-tuned for causal forecasting while keeping the encoder frozen. It retrieves the correct future among 591 candidates in 69.0\% of cases, versus 55.3\% for persistence, and outperforms the compact baseline in latent cosine similarity and retrieval. Gains over persistence are greater at longer horizons and on tokens exhibiting greater change.

These latent gains yield modest L1 improvements but little improvement under LPIPS. Reconstruction error remains even with true future tokens, motivating evaluation through latent and semantic measures alongside pixel metrics and an oracle reconstruction reference.

Semantic analysis indicates driving-relevant information in predicted features. A fixed linear probe trained on real encoder tokens achieves 0.16 mIoU on patches undergoing semantic-class changes, versus 0.13 for persistence. Class-conditioned retrieval also improves over persistence across all six evaluated classes at the longest horizon.

Results are restricted to simulated driving and a prediction horizon of 16 frames, represented by eight temporal groups. The predictor is not conditioned on future ego actions, while object-level forecasting and downstream utility remain unverified.

These findings motivate further analysis of predicted representations for driving-relevant tasks, including future semantic segmentation and object-motion anticipation, to determine whether latent forecasting gains translate into useful perception beyond persistence.

\bibliographystyle{unsrt}
\bibliography{bib} 

\end{document}